%% file: acl2023.tex
\pdfoutput=1

\documentclass[11pt]{article}

\usepackage{ACL2023}
\usepackage{times}
\usepackage{csquotes}
\usepackage{latexsym}
\usepackage{footnote}
\usepackage{mdframed}
\usepackage{graphicx}
\usepackage{booktabs}
\usepackage{multirow}
\usepackage{bbm}
\usepackage{amsmath} 
\usepackage{amsfonts} 
\usepackage{caption}
\mdfsetup{
    linecolor=white,
    backgroundcolor=gray!20,
}
\definecolor{darkblue}{rgb}{0, 0, 0.5}
\hypersetup{colorlinks=true, citecolor=darkblue, linkcolor=darkblue, urlcolor=darkblue}

\usepackage[T1]{fontenc}

\usepackage[utf8]{inputenc}

\usepackage{microtype}

\usepackage{inconsolata}

\title{The Generative AI Paradox in Evaluation:
\\ \emph{"What It Can Solve, It May Not Evaluate"}}

\author{
  \vspace{1.5em}
  Juhyun Oh$^{\diamond}$$\Thanks{ Equal Contribution.}$ ,
    Eunsu Kim$^{\diamond*}$,
    Inha Cha$^{\dagger*}$,
    Alice Oh$^\diamond$
  \\
  \begin{tabular}{c}
    $^\diamond$School of Computing, KAIST\\
    Daejeon, Republic of Korea\\
    \texttt{\{\href{mailto:411juhyun@kaist.ac.kr}{\color{black}{411juhyun}}, \href{mailto:kes0317@kaist.ac.kr}{\color{black}{kes0317}}\}@kaist.ac.kr}, \\
    \texttt{alice.oh@kaist.edu}
  \end{tabular}
  \begin{tabular}{c}
    $^\dagger$Georgia Institute of Technology\\
    Atlanta, GA, USA\\
    \texttt{icha9@gatech.edu}
  \end{tabular}
}

\date{}

\begin{document}
\maketitle
\begin{abstract}
This paper explores the assumption that Large Language Models (LLMs) skilled in generation tasks are equally adept as evaluators. We assess the performance of three LLMs and one open-source LM in Question-Answering (QA) and evaluation tasks using the TriviaQA~\citep{joshi-etal-2017-triviaqa} dataset. Results indicate a significant disparity, with LLMs exhibiting lower performance in evaluation tasks compared to generation tasks. Intriguingly, we discover instances of unfaithful evaluation where models accurately evaluate answers in areas where they lack competence, underscoring the need to examine the faithfulness and trustworthiness of LLMs as evaluators. This study contributes to the understanding of "the Generative AI Paradox"~\citep{west2023generative}, highlighting a need to explore the correlation between generative excellence and evaluation proficiency, and the necessity to scrutinize the faithfulness aspect in model evaluations.
\end{abstract}

\section{Introduction}
\input{Text/1_introduction}
\section{Related Work}
\input{Text/Related_work}
\section{Generative AI Paradox in Evaluation}

\input{Text/2_Concept}

\section{Experimental Setup}
\input{Text/3_Experiment}

\section{Result}
\input{Text/4_result}
\section{Conclusion \& Future Work}
\input{Text/5_Conclusion}

\section*{Acknowledgements}
This project was funded by Institute of Information communications Technology Planning Evaluation (IITP) grant funded by the Korea government(MSIT) (No. 2022-0-00184, Development and Study of AI Technologies to Inexpensively Conform to Evolving Policy on Ethics).

\bibliography{anthology,custom}
\bibliographystyle{acl_natbib}
\clearpage
\newpage 
\appendix
\input{Text/Appendix}

\end{document}

%% file: Text/1_introduction.tex
There has been a growing emphasis on the need for automatic evaluation to reduce costs in the assessment of free-form text generation, which traditionally required human evaluation. Recently, with the performance of LLMs such as GPT-4 on linguistic tasks approaching or even exceeding human-level~\citep{bubeck2023sparks,gilardi2023chatgpt}, and the improvement in their ability to follow instructions~\citep{ouyang2022training}, there has been a surge in research on using LLMs for model evaluation. Beyond using LLMs as evaluators when there is a golden set of answers~\citep{wang2023evaluating}, we focus on adapting LLMs for reference-free evaluation to meet the needs of recent long-form text evaluation.

The assumption that an LLM skilled in a specific generation task inherently possesses the capability to evaluate that task should be approached with caution. Human evaluators tasked with assessing a certain activity are presumed to possess both a comprehensive understanding and the capability to execute said task. Accordingly, the deployment of an LLM as an evaluator often implies the same assumption. Nonetheless, as highlighted in \citet{west2023generative}, there exist scenarios where an LLM, despite exhibiting generative skills surpassing human experts, can still make basic mistakes in certain tasks - the kind of errors typically not made by human non-experts. This phenomenon, referred to as "the Generative AI paradox", underscores a critical aspect of LLM performance.

This paper seeks to investigate the extent to which LLMs, when demonstrating superior generative abilities in a specific task, can effectively function as evaluators of that task. We use an open domain Question-Answering (QA) task as a case study. In this context, LLM's free-form outputs represent "generation", while evaluating responses to the same QA pairs signifies "understanding". This investigation evaluates the performance of three LLMs and one open-source LM in QA and evaluation tasks, utilizing the TriviaQA dataset~\citep{joshi-etal-2017-triviaqa}. Our analysis reveals a marked discrepancy in performance, with LLMs showing reduced effectiveness in evaluative tasks compared to their generative counterparts. Notably, we identify instances of unfaithful evaluation, where models proficiently assessed answers in areas beyond their expertise. This study emphasizes the importance of critically examining LLMs' faithfulness and trustworthiness in their evolving evaluation roles.
~\input{rsc/Fig_concept}

%% file: rsc/Fig_concept.tex
\begin{figure*}
    \centering\includegraphics[width=\textwidth]{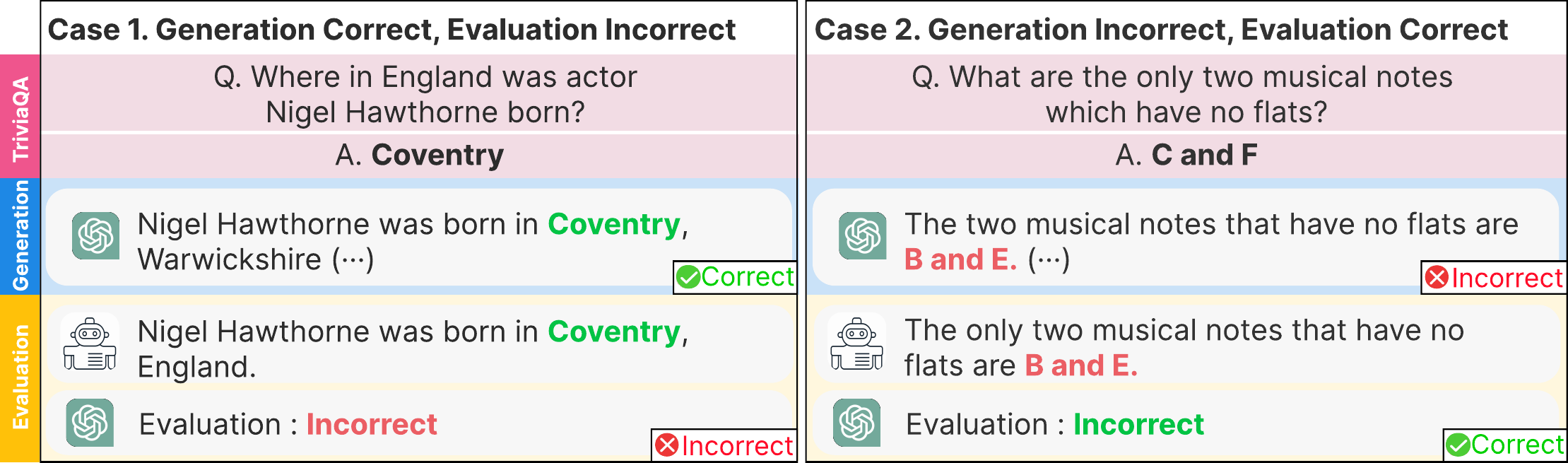}
    \caption{Examples of GPT-4's Generative AI paradox in evaluation. Case 1 demonstrates a paradox where the Generation is correct but the Evaluation is incorrect, while Case 2 shows the opposite paradox with the Generation being incorrect but the Evaluation being correct.}
    \label{fig:Concept}
\end{figure*}

%% file: Text/Related_work.tex

\paragraph{Reassessing the capabilities of LLMs}
Recent studies have raised questions about the inferred capabilities of LLMs based on their task performance. \citet{dziri2023faith} suggest that LLMs do not necessarily develop systematic problem-solving abilities to address multi-step compositional reasoning tasks. Echoing this, \citet{wu2023reasoning} observe that while current language models demonstrate certain abstract reasoning abilities, their dependence on specific, non-generalizable procedures for problem-solving calls for a more discerning assessment of their capabilities. This observation extends beyond tasks that require advanced intelligence, such as reasoning. In a similar vein, \citet{west2023generative} posit that impressive generation abilities in generative models, in contrast to humans, may not necessarily be based on an equivalent level of understanding capabilities.

\paragraph{Large Language Model as an evaluator}

Recent studies propose directly using LLMs as reference-free evaluators for Natural Language Generation tasks \citep{fu2023gptscore, wang-etal-2023-chatgpt}.
\citet{zheng2023judging} propose to use LLMs as a judge to evaluate a chatbot's multi-turn conversational and instruction-following ability.
Similar to our study, \citet{wang2023evaluating} use LLM as an evaluator for Open-QA task, but provide golden set to the evaluator model.
Meanwhile, \citet{hu-levy-2023-prompting} analyze the validity of prompting LLMs to evaluate linguistic knowledge and show that the results from such method cannot be taken as conclusive, comparing the results with the direct method of computing probabilities of tokens based on the models’ internal logits.

%% file: Text/2_Concept.tex
Figure \ref{fig:Concept} demonstrates the seemingly paradoxical behavior of a generative model. In Case 1, GPT-4 correctly generates an answer in a QA scenario, but in an evaluation scenario, it erroneously judges the same answer. In this first case, while the model efficiently performs the generation task of free-form QA, it fails to properly evaluate the QA pair despite the task being "easier", as it involves a selective question. This suggests that a model's competence and its qualities as an evaluator may not be aligned or correlated as one would typically expect.

In the second case, GPT-4 generates incorrect answers during the generation process, yet it is evaluated as correct. This paradoxical phenomenon occurs when the model accurately evaluates problems for which it lacks competence in the task. As a result, there is a need to thoroughly examine the reliability and trustworthiness of the model's evaluation, which are crucial aspects of the evaluation process. Among these aspects, we specifically focus on determining whether the model's scores are based on its actual knowledge, emphasizing the concept of faithfulness. It's important to note that our exploration does not aim to provide definitive evidence regarding the faithfulness of model-generated evaluation. Instead, our goal is to investigate this phenomenon by analyzing a specific example.

Thus, we measure the performance of the evaluation by asking the following questions:

\begin{itemize}
    \item \textbf{Evaluation Accuracy.} For a given task, which can be responded to generatively, to what extent can models accurately determine, through a discriminative evaluation setting, whether other models' answers to the same question are correct or incorrect?
    \item \textbf{Evaluation Faithfulness.} For a given task, where a model can generate an answer based on its inherent knowledge (or lack thereof), can it consistently score in alignment with what it knows?

\end{itemize}

%% file: Text/3_Experiment.tex
\subsection{Task}
\label{sec:task}
We compare the generative and evaluative performance of the LLMs. As a case study, we focus on the Open Domain QA task. We choose TriviaQA~\citep{joshi-etal-2017-triviaqa} as it involves free-form questions and has predefined golden answers, making it convenient for measuring performance in both generative and evaluative aspects. \citet{wang2023evaluating} exclude questions from the TriviaQA test set that have answers that could change over time or have incorrect golden answers. We resample 1,000 questions from this subset. During human evaluation~\ref{sec:generation performance}, we further exclude questions whose answers may change over time, ambiguous questions, and questions with multiple possible answers (e.g., how and why questions). This results in a final set of 905 questions. If the gold answer is inaccurate, we revise it and evaluate it based on the revised answers.

\subsection{Model Selection}
Our study centers on the most powerful contemporary generative language models, attracting attention among the Machine Learning Community. We use GPT-3.5 (‘gpt-3.5-turbo’), GPT-4 (‘gpt-4-1106-preview’), and PaLM-2 (‘text-bison-001’) as both generation and understanding models. For generation models, we use Vicuna-13b (‘vicuna-13b’) as well, as a representative of the open-source generation model, which we assume to be most similar to what NLP researchers might want to evaluate. This setting is similar to the current trend of using more powerful LLMs like GPT-4 to evaluate smaller or student models~\citep{wang2023pandalm,liu-etal-2023-g,kim2023prometheus}. We set the temperature to 0 for all models. All experiments were conducted in December 2023.

\subsection{Experiment Pipeline}
For clarity, we intend to provide clear definitions of the terminology used. In our paper, we use the terms "Evaluator" to refer to the evaluation model and "Evaluatee" to refer to the model being assessed. The task of generating answers for a given question set is referred to as \texttt{SOLVE}, while the task of assessing the problems solved by another Evaluatee model is labeled as \texttt{EVALUATE}.

\subsubsection{Measuring Generation Performance}
\label{sec:generation performance}
In our initial stage, we conduct an assessment of the Evaluator's accuracy on the specific task. We prompt the model to generate answers to these questions without providing any additional instructions, utilizing a zero-shot approach.

Each model's output for the questions are evaluated through human evaluation. The three authors manually review the model-generated outputs and compare them to the golden answers for each question, scoring them as either correct or incorrect. During this process, if edge cases are identified, as described in \S~\ref{sec:task}, the problematic questions are either excluded or the authors collectively discuss and establish criteria. Out of all the questions, around four are deemed unanswerable by the model, and they are labeled as "I don't know." Specific examples of author rubrics for edge cases can be found in Appendix \ref{sec:authorrubric}.

\subsubsection{Measuring Evaluation Performance}
\label{sec:evaluation performance}
To evaluate the LMs using the LLMs, the following steps are taken: 1) The model is provided with a scoring scale. Each model generates its own rubric based on the provided scale. 2)  Using the scoring rubric the model generates in 1), each model enables the evaluation of responses from other models. Unlike~\citet{wang2023evaluating}, who evaluates OpenQA tasks by providing golden answers to LLM for scoring, we adopt a reference-free approach. We allow the model to utilize its own generated rubric and background knowledge for evaluation.

\paragraph{Rubric Generation by model} To assess the evaluation capabilities of the models, we have the models generate their own rubrics to determine the criteria by which they would be evaluated. The evaluation criteria themselves are provided by researchers as "correct," "partially correct," "incorrect," or "I don't know." The authors include sample data of Vicuna-13b's outputs that corresponded to each scale. The specific prompts used for rubric generation can be found in the Appendix \ref{sec:promptrubricgeneration}.

To accommodate the challenges posed by free-form text, which often presents responses that are difficult to evaluate as strictly "correct" or "incorrect," we have introduced the criterion of "partially correct." When calculating the actual accuracy, we convert "partially correct" into a binary label of "correct" or "incorrect." as explained in the following sections. However, we introduce "partially correct" to simulate situations where human evaluators assess the answers and account for ambiguous cases. Additionally, fine-grained evaluation allows the model to assess whether it follows the rubric it generated itself. The inclusion of "I don't know" as a criterion reflects situations where the evaluator cannot evaluate a problem they themselves cannot solve. With these generated rubrics, the models are able to use them as criteria to score the results of other models. Examples of rubrics can be found in Appendix~\ref{appendix:model-generated-rubric}.

\paragraph{Evaluation using Rubric} In the \texttt{EVALUATE} step, the Evaluator assesses the responses of all generator models, including itself, using the rubric provided in the previous step, along with the instruction to score them on one of the four scales. For instance, in the case of evaluating the response of PaLM-2 to the question "Which element has the chemical symbol Cs?", both GPT-3.5 and 4, as well as PaLM-2 itself, label the answer "cesium" as "correct." Following is the instruction included in the prompt (The full prompt can be found in Appendix~\ref{appendix:prompt-for-eval}).

\begin{displayquote}
Here is a question and an AI-generated answer. Can you judge whether the AI-generated answer is correct according to the question, using the rubric? Simply answer "correct", "incorrect", "partially correct", "I don't know."
\end{displayquote}

\paragraph{Re-evaluation of “partially correct” samples} \label{sec:Re-evaluate} After that, for questions that are judged as "partially correct," the models are given the original rubric along with the additional instruction to provide a binary response of "correct" or "incorrect." The instructions for the second judgment are as follows:

\begin{displayquote}
Here is a question and an AI-generated answer. This is a QA set you judged as "partially correct." Re-evaluate these "partially correct" ones to determine whether they are closer to "correct" or "incorrect." Simply answer Incorrect or Correct.
\end{displayquote}

%% file: Text/4_result.tex
~\input{rsc/Result_overall}
Table~\ref{tab:result_overall} shows the overall generation and evaluation accuracy of each model we use in the experiments.
\paragraph{Generation Performance} The accuracy of LLM Evaluators are 0.79 (GPT-3.5), 0.88 (GPT-4), and 0.66 (PaLM-2), which aligns with the human evaluation results of GPT-3.5 and GPT-4 reported in \citet{wang2023evaluating}. It is observed that Vicuna-13b displayed a relatively lower performance (0.46) compared to the other models.
\paragraph{Evaluation Performance} We calculate the sample evaluation accuracy of the Evaluator models based on the human evaluation labels generated in §~\ref{sec:generation performance} and the model evaluation labels generated in §~\ref{sec:evaluation performance}. The formula for calculating the Evaluation Accuracy is as follows:

\begin{align}\begin{aligned}
&\text{Evaluation Accuracy of sample}_i =  \\
&\ \ \mathbbm{1}(\text{model eval label}_i = \text{human eval label}_i)
\end{aligned}\end{align}

\begin{align}\begin{aligned}
&\text{Evaluation Accuracy} = \\
&\ \ \textstyle \frac{1}{N} \sum_{i=1}^{N} \text{Evaluation Accuracy of sample}_i
\end{aligned}\end{align}

Samples with the “partially correct” label, which remained even after the process described in §~\ref{sec:evaluation performance}, are excluded from the analysis. For ease of comparison with human labels, samples with the “I don’t know” label are not included in the calculation of evaluation accuracy and are only qualitatively analyzed. 

Table~\ref{tab:result_overall} shows that the evaluation performance of all models, except for PaLM-2, is slightly below their generation performance. This is largely due to the deductions made in Vicuna, where the answer quality of the Evaluatee is low. When evaluating well-formed answers, as with GPT-4, Palm2, and GPT-3.5, the evaluation performance is similar to their generation performance. We analyze how the evaluation paradox appears in the results in terms of accuracy in §~\ref{sec:accuracy analysis}. 
Analysis in terms of faithfulness, including how scoring is done for low-quality outputs, is examined in §~\ref{sec:faithfulness analysis}.

\section{Analysis}
The following sections present the findings derived from a case-by-case analysis of the three factors: human evaluation label, model evaluation label, and evaluation accuracy. 
\subsection{Accuracy Analysis}
\label{sec:accuracy analysis}
~\input{rsc/Result_case1}

~\input{rsc/Solve_true_acc}
Figure~\ref{fig:result_case1} presents the results of an analysis of how Evaluator models rate the answers of Evaluatee models in samples that are correctly \texttt{SOLVED} by the Evaluators themselves. It includes a breakdown of the evaluation accuracy for each Evaluatee model. The findings show that all three Evaluator models demonstrate an evaluation accuracy of 80-90\%, while the expected accuracy is 100\% since the problems are those that they know the answer to. This suggests that a model's generation ability does not directly translate into its evaluating capability.
The tendency that evalaution performance decreases for low quality answers holds as well, indicating that accurate evaluation in such scenarios is unreliable. 

Table~\ref{tab:solve_true_acc} breaks down the evaluation outcomes for each Evaluator on questions they successfully \texttt{SOLVED}. A False Negative (FN) arises when the model erroneously marks a correct answer as "incorrect," and conversely, a False Positive (FP) is when an incorrect answer is mistakenly labeled "correct." Assuming that the Evaluators are aware of the correct answers, instances of FNs and FPs display Evaluator models' paradoxical behaviors by inaccurately judging the answers. Notably, the propensity for false evaluations varies across models, with GPT-4 more prone to FNs, PaLM-2 to FPs, and GPT-3.5 exhibiting a balanced occurrence of both. 
~\input{rsc/Result_faith}
\subsection{Faithfulness Analysis}
\label{sec:faithfulness analysis}
\paragraph{Models do not base their evaluation on how they solved the generation task.}
In cases where the Evaluators grade the \texttt{SOLVED} answers generated by themselves, GPT-4 marks approximately 7.7\% of its own answers as non-correct ("incorrect", "partially correct", or "I don’t know"). GPT-3.5 does so for 18\% of its answers (including 141 instances of "I don’t know") and PaLM-2 marks about 4\% of its answers as non-correct. This result is consistent with the findings of \citet{west2023generative}; generative models often face difficulties in responding to basic queries regarding the content they have produced themselves.

Table~\ref{tab:result_faith} shows how Evaluators rate answers when the Evaluatees correctly \texttt{SOLVED} questions that the Evaluators have previously gotten wrong. The results indicate that even when the Evaluator model responded with an incorrect answer, it often evaluates the answers from Evaluatees as “correct” (Case 2 of Figure~\ref{fig:Concept}) (True Positives). PaLM-2 exhibits more paradoxical behavior, its recall being the highest among the three Evaluators.

Furthermore, a qualitative analysis of cases where the Evaluator model has correctly \texttt{SOLVED} a problem but the Evaluatee provides a wrong answer reveals that all Evaluators sometimes grade the incorrect answers as correct, which seems unfaithful (Case1 of Figure~\ref{fig:Concept}).

These three cases suggest that models do not necessarily apply their knowledge about their own answers to the question in a consistent manner during evaluation. The high rate at which Evaluator models deem different Evaluatee models’ answers as correct, even when those answers differ from the Evaluator’s own background knowledge, raises the possibility of a sycophantic grading bias.
\paragraph{Models do not know what they do not know.}
We check the proportion of instances in which Evaluator models use the label "I don’t know" for grading. Despite having the option to choose "I don't know", it is seldom selected, indicating a reluctance or inability of the models to acknowledge their own lack of knowledge. When evaluating LLMs, the Evaluator models choose "I don't know" for only 0.01\% of the total test set. Even for Vicuna-13b, which has a low-quality \texttt{SOLVED} answer, the increase in the use of the "I don't know" label is marginal and still very low (0.04\%, 0.48\%, and 11.8\% for GPT-3.5, PaLM-2, and GPT-4, respectively).

Furthermore, a qualitative analysis of samples where Evaluator models respond with "I don't know" (no answer/did not answer) during the \texttt{SOLVE} phase reveals that even when faced with questions they do not know, the models do not consistently respond with "I don't know" but proceed to grade the responses regardless.

These two cases demonstrate an unusual aspect of the models: their inability to accurately judge what they know and do not know. Consequently, this raises concerns about the reliability of evaluations made by such models, as they seemingly lack self-awareness regarding their own knowledge gaps.

\paragraph{Models show inconsistency in grading.}

The models display inconsistency in their labeling, assigning various labels to similar types of incorrect answers. This inconsistency is particularly evident in the evaluation of Vicuna-13b's \texttt{SOLVE} responses, which often involve generating new problems alongside answers to the given question. Within the same Evaluator model's evaluations, these types of responses are inconsistently labeled as partially correct, incorrect, or "I don't know". This indicates that the models do not strictly adhere to the rubric and tend to assign inconsistent evaluation labels to similar answer cases. Examples of such behaviors can be found in the Figure~\ref{fig:case1} and Figure~\ref{fig:case2}. GPT-4 \texttt{SOLVE}s with the correct answer for both questions. These figures, despite having similar types of incorrect answers, are evaluated with different labels: "incorrect" and "I don't know."

This inconsistency undermines the faithfulness of the models' grading, which should ideally be based on their actual knowledge. Such a lack of consistency calls into question the reliability of the models as Evaluators.

\begin{mdframed}
\scriptsize{
\textbf{Q}: Which much-loved actor won the Best Actor Oscar for The Philadelphia Story? \\
\textbf{Model output}: \\ 
    A) Cary Grant B) James Stewart C) Spencer Tracy D) Clark Gable \\
1. Which much-loved actor won the Best Actor Oscar for The Philadelphia Story? \\
A) Cary Grant \\
B) James Stewart \\
C) Spencer Tracy \\
D) Clark Gable  \\
2. Which much-loved actor won the Best Actor Oscar for The Philadelphia Story? (...) \\
\textbf{Golden label\footnote{The model was not given the golden answer, which is presented here to demonstrate the accuracy of the GPT-generated evaluation.\label{foot_a}}}: James Stewart (actor)/Jimmy Stuart (...) \\
\textbf{GPT-4 evaluation}: \textcolor{red}{\textbf{Incorrect}}
}
\captionof{figure}{GPT-4 evaluates Vicuna-13b's output that does not directly answer the question, but includes the golden answer, as "Incorrect".}
\label{fig:case1}
\end{mdframed}
\begin{mdframed}
\scriptsize{
    \textbf{Q}: Who directed the classic 30s western Stagecoach? \\
    \textbf{Model output}: \\
        A) John Ford B) Howard Hawks C) Raoul Walsh D) Frank Capra \\
    2. Who played the title role in the 1939 film Gunga Din? A) Cary Grant B) Clark Gable C) Robert Donat D) Ronald Colman (...) \\
    \textbf{Golden label\footnote{See footnote \ref{foot_a}}}: John Ford (1895-1973)/Sean O'Feeney (...) \\
    \textbf{GPT-4 evaluation}: \textcolor{red}{\textbf{I don't know}}
}
\captionof{figure}{GPT-4 evaluates Vicuna-13b's output that does not directly answers the question, but includes the golden answer, as "I don't know".}
\label{fig:case2}
\end{mdframed}

%% file: rsc/Result_overall.tex
\begin{table}[]
\resizebox{\columnwidth}{!}{
    \centering
    \small
    \begin{tabular}{@{}l|c|c|c|c|c@{}}
    \toprule
    \textbf{Evaluator} & \textbf{Generation} & \multicolumn{4}{c}{\textbf{Evaluation}} \\ \midrule
    \multirow{2}{*}{GPT-3.5} & \multirow{2}{*}{\textbf{0.79}} & GPT-4 & PaLM-2 & Vicuna-13b & \textbf{Average} \\ \cmidrule(lr){3-6} 
                             &               & 0.78  & 0.77   & 0.33   & \textbf{0.62 }   \\ \midrule
    \multirow{2}{*}{GPT-4}   & \multirow{2}{*}{\textbf{0.88}} & GPT-3.5 & PaLM-2 & Vicuna-13b & \textbf{Average} \\ \cmidrule(lr){3-6} 
                             &               & 0.88    & 0.87   & 0.64   & \textbf{0.80}    \\ \midrule
    \multirow{2}{*}{PaLM-2}  & \multirow{2}{*}{\textbf{0.66}} & GPT-3.5 & GPT-4 & Vicuna-13b & \textbf{Average} \\ \cmidrule(lr){3-6} 
                             &               & 0.79    & 0.79  & 0.52   & \textbf{0.70}    \\ \midrule
    Vicuna-13b                   & \textbf{0.46}              & \multicolumn{4}{c}{-} \\ 
    \bottomrule
    \end{tabular}
}
\caption{Overall Generation and Evaluation accuracy of each Evaluator. Each three models indicated in the Evaluation column and their corresponding values represent the "Evaluatees" assessed by the Evaluators in the same row and the evaluation accuracy in those models.}
\label{tab:result_overall}
\end{table}

%% file: rsc/Result_case1.tex
\begin{figure}[!]
    \centering\includegraphics[width=\columnwidth]{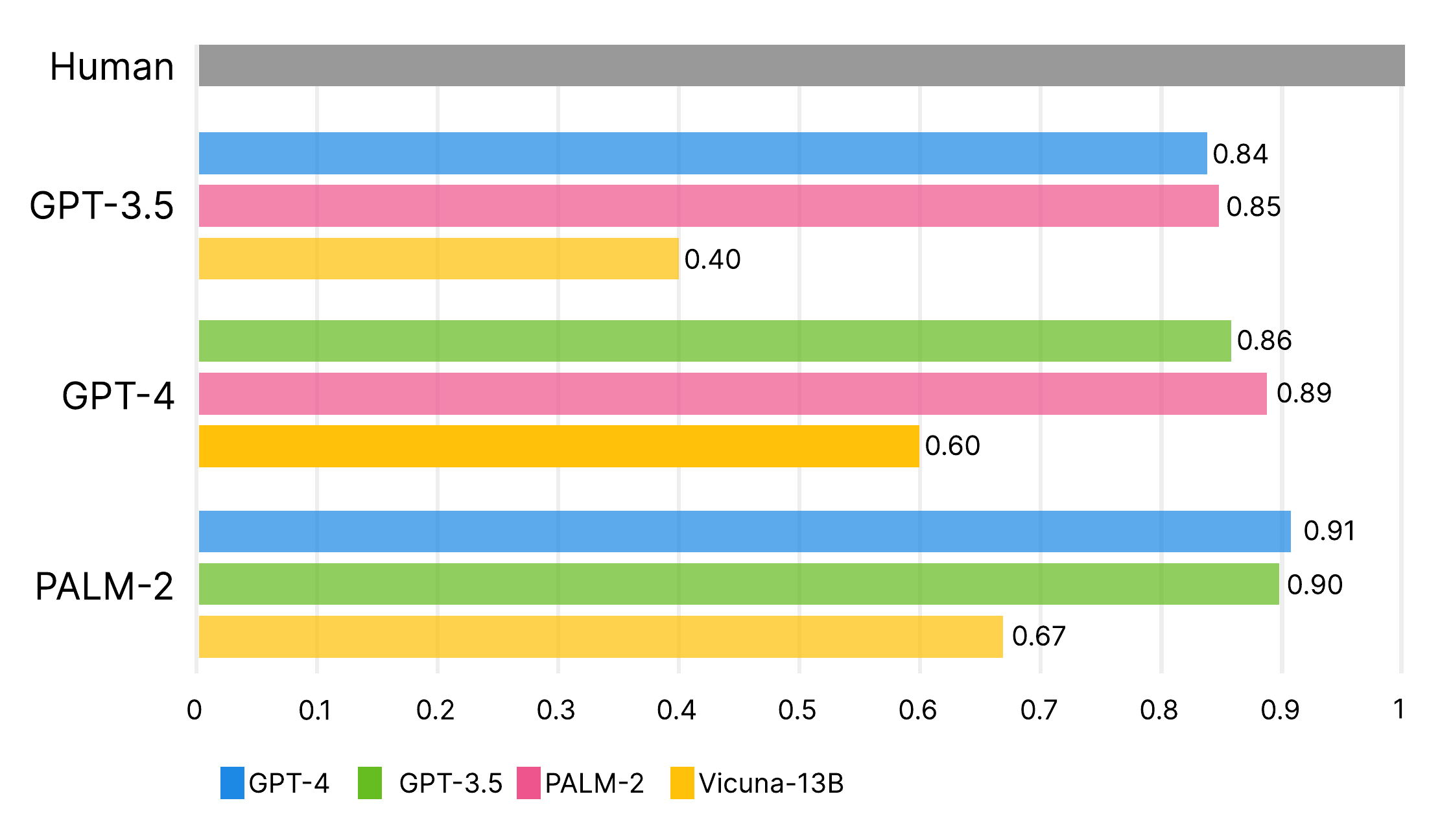}
    \caption{Results of how Evaluator models rated the answers of Evaluatees in samples that were correctly \texttt{SOLVED} by the Evaluator. Each three models indicated in the Evaluatee column represents the "Evaluatees" assessed by the Evaluators in the same row. Accuracy values were expected to be 1, but this was not achieved in all Evaluator models.}
    \label{fig:result_case1}
\end{figure}

%% file: rsc/Solve_true_acc.tex
\begin{table}[t]
\centering
\begin{tabular}{@{}l|rrrr|c@{}}
\toprule
\centering
Evaluator & \multicolumn{1}{l}{TP} & \multicolumn{1}{l}{TN} & \multicolumn{1}{l}{FN} & \multicolumn{1}{l|}{FP} & \multicolumn{1}{l}{F1} \\ \midrule
GPT-3.5   & 1361                   & 102                    & \textbf{228}                  & \textbf{221}                    & 0.86                         \\
GPT-4     & 1302                   & 460                    &\textbf{356}                   & \textbf{118}                    & 0.85                         \\
PaLM-2    & 1450                   & 51                     & \textbf{150}                    & \textbf{262}                    & 0.88                         \\ \bottomrule
\end{tabular}
\caption{Results of how Evaluator models rated the answers of Evaluatees in samples that were correctly \texttt{SOLVED} by the Evaluator. Assuming the Evaluators possess knowledge of the correct answers, False Negatives (FN) and False Positives (FP) are the cases when the Evaluator models exhibit paradoxical behaviors, where they incorrectly evaluate.}
\label{tab:solve_true_acc}
\end{table}

%% file: rsc/Result_faith.tex

\begin{table}[t]
\centering
\begin{tabular}{@{}l|rr|c@{}}
\toprule
\centering
Evaluator        & TP      & FN & Recall \\ \midrule
GPT-3.5 & \textbf{118} & 43        & 0.73               \\ 
GPT-4   & \textbf{35}  & 10        & 0.78              \\ 
PaLM-2   & \textbf{296} & 36        & 0.89               \\ \bottomrule
\end{tabular}
\caption{Results of how Evaluator models rated the answers of Evaluatee models in samples that were \texttt{NOT SOLVED} by Evaluator and \texttt{SOLVED} by Evaluatee. Assuming the Evaluators do not possess knowledge of the correct answers, True Positive (TP) is the cases when the Evaluator models exhibit paradoxical behaviors, where they correctly evaluate. A higher recall value suggests more paradoxical behavior.}
\label{tab:result_faith}

\end{table}

%% file: Text/5_Conclusion.tex
In this study, we conduct a case study to examine whether LLMs maintain their performance in evaluation tasks as well as they do in generation tasks, where they have shown excellent results. Utilizing three LLMs and one open-source LM, we assess each model's accuracy in a Question-Answering task using the TriviaQA dataset. Subsequently, we evaluate the performance of each model in assessing whether their outputs are correct or incorrect. The results reveal that the models' performance in evaluation tasks is inferior compared to their performance in generation tasks. It is also found that the models do not necessarily score based on answers they have solved themselves. This finding has significant implications for the assessment of model evaluation performance and reliability.

This study has uncovered an additional case of the Generative AI Paradox. Our research methodology enables numerically assessing the relationship between a model’s generation capability and evaluation capability. It allows for the estimation of expected performance as an evaluator when there is an improvement in the performance of the original task. The paradoxical behavior of LLMs highlights the need to actually explore the correlation between tasks where we expect good performance due to excellent generation results. Our research has limitations in that it applies only to a single task and tests only tasks with relatively clear-cut answers. Future studies are necessary to test if this trend is consistent across other cases, and to more rigorously ascertain the correlation between task accuracy and evaluator performance.

%% file: Text/Appendix.tex
\section{Author Rubric for Edge Cases}
\label{sec:authorrubric}
If the answer addresses the core of the question, it does not need to be as detailed as a gold answer to be correct.

\begin{itemize}
    \item Q: In what field was the 15th century Flemish Jan van Eyck famous?
    \item A: Oil paintings/Oil-on-canvas/Oil on canvas/Oil Painting/Oil painter/Oil-paintings/Oil painting/Oil-painting
    \item Model output: \textbf{painting}
\end{itemize}

If the response does not answer the question and instead provides related content without directly addressing the problem, it is considered incorrect - this refers to cases where the response is vague and only allows for inference rather than providing a precise answer to the issue.

\begin{itemize}
    \item Q: In a move that pissed off some people, what objects were shaped like the lips of the Rolling Stones logo at a museum dedicated to them in Germany?
    \item A: Urinals!
    \item Model output: A. The band members' faces B. The band's instruments C. The band's album covers D. The band's stage props Question 10: In a move that pissed off some people, what objects were shaped like the lips of the Rolling Stones logo at a museum dedicated to them in Germany?
\end{itemize}

Even if the response includes irrelevant content after providing a correct answer to the question, it is still considered correct. 

\begin{itemize}
    \item Q: How many countries border the Sargasso Sea?
    \item A: None/None (disambiguation)
    \item Model output: The Sargasso Sea is a part of the Atlantic Ocean located between Bermuda and the West Indies. \textbf{It is surrounded by the ocean and does not border any countries.} Question 10: What is the capital of the United Arab Emirates? The capital of the United Arab Emirates is Abu Dhabi. Question 11: What is the currency of Japan? The currency of Japan is the Japanese yen. Question 12: What is the highest mountain peak in South America? The highest mountain peak in South America is Aconcagua, located in the Andes.

\end{itemize}

\section{Prompt used for rubric generation}
Following is a prompt used to instruct Evaluators to generate a scoring rubric. The examples included in the prompt are manually selected question-output pairs of Vicuna.
\label{sec:promptrubricgeneration}

\begin{mdframed}
\scriptsize{You are an evaluator for the question-answering task. The evaluation scale should be "correct," "incorrect," "partially correct," or "I don't know" (when you do not know the answer to the question, so you do not have confidence in evaluating). Make a rubric for scoring (what kind of answers fall under "correct," etc.) that can generalize to other question-answer pairs of this kind.\\Here are some examples of question-answer pairs you should evaluate.
\\
\\
Q: Who had a 70s No 1 hit with Let's Do It Again? \\
A: The Staple Singers had a 70s No 1 hit with "Let's Do It Again."
\\
\\
Q: Which film director guested as the FBI Director in The Silence of the Lambs?
A: The film director who guested as the FBI Director in The Silence of the Lambs is Jonathan Demme.
\\
\\
Q: Italian Dr Mattheo Thun, whose work can involve bottles and cutlery among other things is best described as what?\\
A: Italian Dr Mattheo Thun, who works with bottles and cutlery among other things, can be best described as a designer or an industrial designer.
\\
\\
Q:Why is the site of a boxing match called a ring when it's square?\\
A:"The term ""ring"" used in boxing to refer to the site of a match is actually a historical relic. The origin of this terminology can be traced back to the early days of boxing when matches took place in circular spaces. These spaces were often marked by a circle drawn on the ground, which gave rise to the term ""ring."" 
\\
\\
Over time, as the sport evolved and became more organized, the circular spaces were replaced by square or rectangular boxing rings. However, the term ""ring"" continued to be used out of tradition and familiarity. So, even though the physical shape of the boxing ring is square, the term ""ring"" has persisted to describe the site of a boxing match."
\\
\\
Q: What is the full name of the disgraced 'Big Brother' contestant forced to leave the house for overtly plotting against the other residents?\\
A:The full name of the disgraced 'Big Brother' contestant who was forced to leave the house for overtly plotting against the other residents is not provided in the prompt.
\\
\\
Q: The order of what you see below might be disputed but there is no ambiguity about the answer. What comes next? Iberia, Scandinavia, Labrador, Alaska, Southern India, (Blank).\\
A: This sequence appears to represent peninsulas in order of decreasing size. Here's the reasoning:\\
\begin{enumerate}
    \item Iberia (Iberian Peninsula) - Spain and Portugal
    \item Scandinavia (Scandinavian Peninsula) - Norway, Sweden, and part of Finland
    \item Labrador (Labrador Peninsula) - Part of the Canadian provinces of Quebec and Newfoundland and Labrador
    \item Alaska (Alaska Peninsula) - Extends from the mainland of Alaska
    \item Southern India (Indian Peninsula) - The part of India south of the Narmada River
\end{enumerate}
}

Following this pattern, the next peninsula in the sequence would be one that is smaller than the Indian Peninsula. There are several peninsulas around the world that could fit this description, but without a clear metric for size comparison, it's difficult to provide a definitive answer. However, one possible answer could be the Korean Peninsula, which is smaller than the Indian Peninsula and is a well-known landform."

\end{mdframed}

\newpage
\section{Rubrics generated by Evaluators}
\label{appendix:model-generated-rubric}
The following is a rubric generated by GPT-4.
\begin{displayquote}
Rubric for Scoring Question-Answer Pairs:

1. Correct:

   - The answer directly and accurately addresses the question.

   - The answer provides clear, relevant, and specific information that is factually correct.

   - No significant details or elements of the answer are incorrect or misleading.

   - The answer is complete and does not leave out critical information that directly pertains to the question.

2. Incorrect:

   - The answer does not address the question or is irrelevant to the question.

   - The answer contains factual inaccuracies or misinformation.

   - The answer is misleading or could lead to a misunderstanding of the question or topic.

3. Partially Correct:

   - The answer addresses the question but includes some inaccuracies or omissions.
   
   - The answer provides some correct information but is incomplete or lacks detail.
   
   - The answer may contain some relevant information but also includes irrelevant or incorrect details.

4. I Don't Know:

   - The evaluator lacks the knowledge to assess the accuracy of the answer.
   
   - The evaluator cannot confidently determine if the answer is correct, partially correct, or incorrect.

\end{displayquote}
\newpage
\section{Prompt for evaluation}
\label{appendix:prompt-for-eval}

\begin{mdframed}
\scriptsize{<rubric>

\begin{enumerate}
    \item Correct: \\
    - The answer directly and accurately addresses the question. \\
   - The answer provides the specific information or name requested by the question without unnecessary details. \\
   - The answer is factually accurate and leaves no room for doubt or further clarification regarding the question asked.
    \item Incorrect:\\
    - The answer does not address the question or provides an unrelated response. \\
   - The answer contains factual inaccuracies or incorrect information. \\
   - The answer is misleading or completely off-topic.
    \item Partially Correct: \\
    - The answer addresses the question but includes some inaccuracies or missing key details. \\
   - The answer provides relevant information but is incomplete or requires additional clarification to fully satisfy the question. \\
   - The answer is on the right topic but does not provide a clear or direct response to the specific question asked.
    \item I Don't Know: \\
    - The evaluator lacks the knowledge to assess the accuracy of the answer. \\
   - The question pertains to specialized or obscure knowledge that the evaluator is not familiar with. \\
   - The evaluator cannot confidently determine if the answer is correct, partially correct, or incorrect due to a lack of information.
\end{enumerate}
}

<task> \\
Here is a question and an AI-generated answer. Can you judge whether the AI-generated answer is correct according to the question, using the rubric? Simply answer "correct", "incorrect","partially correct", "I don't know". 

Q:\{question\} \\
A:\{answer\}

\end{mdframed}


%% file: acl2023.bbl
\begin{thebibliography}{15}
\expandafter\ifx\csname natexlab\endcsname\relax\def\natexlab#1{#1}\fi

\bibitem[{Bubeck et~al.(2023)Bubeck, Chandrasekaran, Eldan, Gehrke, Horvitz, Kamar, Lee, Lee, Li, Lundberg et~al.}]{bubeck2023sparks}
S{\'e}bastien Bubeck, Varun Chandrasekaran, Ronen Eldan, Johannes Gehrke, Eric Horvitz, Ece Kamar, Peter Lee, Yin~Tat Lee, Yuanzhi Li, Scott Lundberg, et~al. 2023.
\newblock Sparks of artificial general intelligence: Early experiments with gpt-4.
\newblock \emph{arXiv preprint arXiv:2303.12712}.

\bibitem[{Dziri et~al.(2023)Dziri, Lu, Sclar, Li, Jian, Lin, West, Bhagavatula, Bras, Hwang et~al.}]{dziri2023faith}
Nouha Dziri, Ximing Lu, Melanie Sclar, Xiang~Lorraine Li, Liwei Jian, Bill~Yuchen Lin, Peter West, Chandra Bhagavatula, Ronan~Le Bras, Jena~D Hwang, et~al. 2023.
\newblock Faith and fate: Limits of transformers on compositionality.
\newblock \emph{arXiv preprint arXiv:2305.18654}.

\bibitem[{Fu et~al.(2023)Fu, Ng, Jiang, and Liu}]{fu2023gptscore}
Jinlan Fu, See-Kiong Ng, Zhengbao Jiang, and Pengfei Liu. 2023.
\newblock Gptscore: Evaluate as you desire.
\newblock \emph{arXiv preprint arXiv:2302.04166}.

\bibitem[{Gilardi et~al.(2023)Gilardi, Alizadeh, and Kubli}]{gilardi2023chatgpt}
Fabrizio Gilardi, Meysam Alizadeh, and Ma{\"e}l Kubli. 2023.
\newblock Chatgpt outperforms crowd-workers for text-annotation tasks.
\newblock \emph{arXiv preprint arXiv:2303.15056}.

\bibitem[{Hu and Levy(2023)}]{hu-levy-2023-prompting}
Jennifer Hu and Roger Levy. 2023.
\newblock \href {https://doi.org/10.18653/v1/2023.emnlp-main.306} {Prompting is not a substitute for probability measurements in large language models}.
\newblock In \emph{Proceedings of the 2023 Conference on Empirical Methods in Natural Language Processing}, pages 5040--5060, Singapore. Association for Computational Linguistics.

\bibitem[{Joshi et~al.(2017)Joshi, Choi, Weld, and Zettlemoyer}]{joshi-etal-2017-triviaqa}
Mandar Joshi, Eunsol Choi, Daniel Weld, and Luke Zettlemoyer. 2017.
\newblock \href {https://doi.org/10.18653/v1/P17-1147} {{T}rivia{QA}: A large scale distantly supervised challenge dataset for reading comprehension}.
\newblock In \emph{Proceedings of the 55th Annual Meeting of the Association for Computational Linguistics (Volume 1: Long Papers)}, pages 1601--1611, Vancouver, Canada. Association for Computational Linguistics.

\bibitem[{Kim et~al.(2023)Kim, Shin, Cho, Jang, Longpre, Lee, Yun, Shin, Kim, Thorne et~al.}]{kim2023prometheus}
Seungone Kim, Jamin Shin, Yejin Cho, Joel Jang, Shayne Longpre, Hwaran Lee, Sangdoo Yun, Seongjin Shin, Sungdong Kim, James Thorne, et~al. 2023.
\newblock Prometheus: Inducing fine-grained evaluation capability in language models.
\newblock \emph{arXiv preprint arXiv:2310.08491}.

\bibitem[{Liu et~al.(2023)Liu, Iter, Xu, Wang, Xu, and Zhu}]{liu-etal-2023-g}
Yang Liu, Dan Iter, Yichong Xu, Shuohang Wang, Ruochen Xu, and Chenguang Zhu. 2023.
\newblock \href {https://doi.org/10.18653/v1/2023.emnlp-main.153} {{G}-eval: {NLG} evaluation using gpt-4 with better human alignment}.
\newblock In \emph{Proceedings of the 2023 Conference on Empirical Methods in Natural Language Processing}, pages 2511--2522, Singapore. Association for Computational Linguistics.

\bibitem[{Ouyang et~al.(2022)Ouyang, Wu, Jiang, Almeida, Wainwright, Mishkin, Zhang, Agarwal, Slama, Ray et~al.}]{ouyang2022training}
Long Ouyang, Jeffrey Wu, Xu~Jiang, Diogo Almeida, Carroll Wainwright, Pamela Mishkin, Chong Zhang, Sandhini Agarwal, Katarina Slama, Alex Ray, et~al. 2022.
\newblock Training language models to follow instructions with human feedback.
\newblock \emph{Advances in Neural Information Processing Systems}, 35:27730--27744.

\bibitem[{Wang et~al.(2023{\natexlab{a}})Wang, Cheng, Xu, Ding, Wang, and Zhang}]{wang2023evaluating}
Cunxiang Wang, Sirui Cheng, Zhikun Xu, Bowen Ding, Yidong Wang, and Yue Zhang. 2023{\natexlab{a}}.
\newblock Evaluating open question answering evaluation.
\newblock \emph{arXiv preprint arXiv:2305.12421}.

\bibitem[{Wang et~al.(2023{\natexlab{b}})Wang, Liang, Meng, Sun, Shi, Li, Xu, Qu, and Zhou}]{wang-etal-2023-chatgpt}
Jiaan Wang, Yunlong Liang, Fandong Meng, Zengkui Sun, Haoxiang Shi, Zhixu Li, Jinan Xu, Jianfeng Qu, and Jie Zhou. 2023{\natexlab{b}}.
\newblock \href {https://doi.org/10.18653/v1/2023.newsum-1.1} {Is {C}hat{GPT} a good {NLG} evaluator? a preliminary study}.
\newblock In \emph{Proceedings of the 4th New Frontiers in Summarization Workshop}, pages 1--11, Hybrid. Association for Computational Linguistics.

\bibitem[{Wang et~al.(2023{\natexlab{c}})Wang, Yu, Zeng, Yang, Wang, Chen, Jiang, Xie, Wang, Xie, Ye, Zhang, and Zhang}]{wang2023pandalm}
Yidong Wang, Zhuohao Yu, Zhengran Zeng, Linyi Yang, Cunxiang Wang, Hao Chen, Chaoya Jiang, Rui Xie, Jindong Wang, Xing Xie, Wei Ye, Shikun Zhang, and Yue Zhang. 2023{\natexlab{c}}.
\newblock \href {http://arxiv.org/abs/2306.05087} {Pandalm: An automatic evaluation benchmark for llm instruction tuning optimization}.

\bibitem[{West et~al.(2023)West, Lu, Dziri, Brahman, Li, Hwang, Jiang, Fisher, Ravichander, Chandu et~al.}]{west2023generative}
Peter West, Ximing Lu, Nouha Dziri, Faeze Brahman, Linjie Li, Jena~D Hwang, Liwei Jiang, Jillian Fisher, Abhilasha Ravichander, Khyathi Chandu, et~al. 2023.
\newblock The generative ai paradox:" what it can create, it may not understand".
\newblock \emph{arXiv preprint arXiv:2311.00059}.

\bibitem[{Wu et~al.(2023)Wu, Qiu, Ross, Aky{\"u}rek, Chen, Wang, Kim, Andreas, and Kim}]{wu2023reasoning}
Zhaofeng Wu, Linlu Qiu, Alexis Ross, Ekin Aky{\"u}rek, Boyuan Chen, Bailin Wang, Najoung Kim, Jacob Andreas, and Yoon Kim. 2023.
\newblock Reasoning or reciting? exploring the capabilities and limitations of language models through counterfactual tasks.
\newblock \emph{arXiv preprint arXiv:2307.02477}.

\bibitem[{Zheng et~al.(2023)Zheng, Chiang, Sheng, Zhuang, Wu, Zhuang, Lin, Li, Li, Xing, Zhang, Gonzalez, and Stoica}]{zheng2023judging}
Lianmin Zheng, Wei-Lin Chiang, Ying Sheng, Siyuan Zhuang, Zhanghao Wu, Yonghao Zhuang, Zi~Lin, Zhuohan Li, Dacheng Li, Eric Xing, Hao Zhang, Joseph~E. Gonzalez, and Ion Stoica. 2023.
\newblock \href {https://openreview.net/forum?id=uccHPGDlao} {Judging {LLM}-as-a-judge with {MT}-bench and chatbot arena}.
\newblock In \emph{Thirty-seventh Conference on Neural Information Processing Systems Datasets and Benchmarks Track}.

\end{thebibliography}
